\documentclass{article}

\usepackage[pagebackref,breaklinks,colorlinks,citecolor=cyan,linkcolor=cyan]{hyperref}

\usepackage[preprint]{neurips_2025}

\usepackage[utf8]{inputenc} %
\usepackage[T1]{fontenc}    %
\usepackage{hyperref}       %
\usepackage{url}            %
\usepackage{booktabs}       %
\usepackage{graphicx}       %
\usepackage{amsmath}        %
\usepackage[table]{xcolor} %
\usepackage{amsfonts}       %
\usepackage{nicefrac}       %
\usepackage{microtype}      %
\usepackage{xcolor}         %
\usepackage{multirow}
 \usepackage{colortbl}
 \usepackage{makecell}
\usepackage{wrapfig}
\usepackage{float}
\RequirePackage{subcaption}
\usepackage{enumitem}

\usepackage{soul} %
\usepackage{todonotes} %

\definecolor{ReviewRed}{RGB}{255,69,0} %

\definecolor{Magenta}{rgb}{0.8, 0.1, 0.6}

\title{DiffusionReward: Enhancing Blind Face Restoration through Reward Feedback Learning}

\author{%
  Bin Wu\textsuperscript{1,2}, %
  Wei Wang\textsuperscript{1,2}\thanks{Corresponding author.}, %
  Yahui Liu\textsuperscript{3}, %
  Zixiang Li\textsuperscript{1,2}, %
  Yao Zhao\textsuperscript{1,2} \\ %
  \textsuperscript{1}Institute of Information Science, Beijing Jiaotong University \\
  \textsuperscript{2}Visual Intelligence +X International Cooperation Joint Laboratory of MOE \\
  \textsuperscript{3}Kuaishou Technology
}

\usepackage{caption}
\begin{document}

\maketitle

\begin{abstract}
Reward Feedback Learning (ReFL) has recently shown great potential in aligning model outputs with human preferences across various generative tasks.
In this work, we introduce a ReFL framework, named \textit{DiffusionReward}, to the Blind Face Restoration task for the first time.
DiffusionReward effectively overcomes the limitations of diffusion-based methods, which often fail to generate realistic facial details and exhibit poor identity consistency. 
The core of our framework is the Face Reward Model (FRM), which is trained using carefully annotated data.
It provides feedback signals that play a pivotal role in steering the optimization process of the restoration network.
In particular, our ReFL framework incorporates a gradient flow into the denoising process of \textit{off-the-shelf} face restoration methods to guide the update of model parameters.
The guiding gradient is collaboratively determined by three aspects: (i) the FRM to ensure the perceptual quality of the restored faces; (ii) a regularization term that functions as a safeguard to preserve generative diversity; and (iii) a structural consistency constraint to maintain facial fidelity. 
Furthermore, the FRM undergoes dynamic optimization throughout the process. It not only ensures that the restoration network stays precisely aligned with the real face manifold, but also effectively prevents reward hacking.
Experiments on synthetic and wild datasets demonstrate that our method outperforms state-of-the-art methods, significantly improving identity consistency and facial details. %
The source codes, data and models are available at: \href{https://github.com/01NeuralNinja/DiffusionReward}{https://github.com/01NeuralNinja/DiffusionReward}.
\end{abstract}

\section{Introduction}
\label{sec:introduction}

Facial images captured in-the-wild often suffer from complex and diverse degradations, such as blur, compression artifacts, noise, and low resolution. Blind Face Restoration (BFR)~\cite{li2018learning,li2020blind,wang2021towards} aims to restore high-quality (HQ) counterparts from these degraded inputs. 
Given the substantial information loss in low-quality (LQ) inputs and the typically unknown degradation processes, BFR is inherently a highly ill-posed problem.
As a result, for any given single LQ face, there theoretically exists a solution space encompassing an infinite number of potential high-quality solutions. 
Consequently, accurately reconstructing HQ facial images from this expansive solution space 
remains an unsolved challenge, especially in terms of photorealism, naturalness, and identity preservation.

\begin{figure}[t]
    \centering
    \includegraphics[width=0.9\columnwidth]{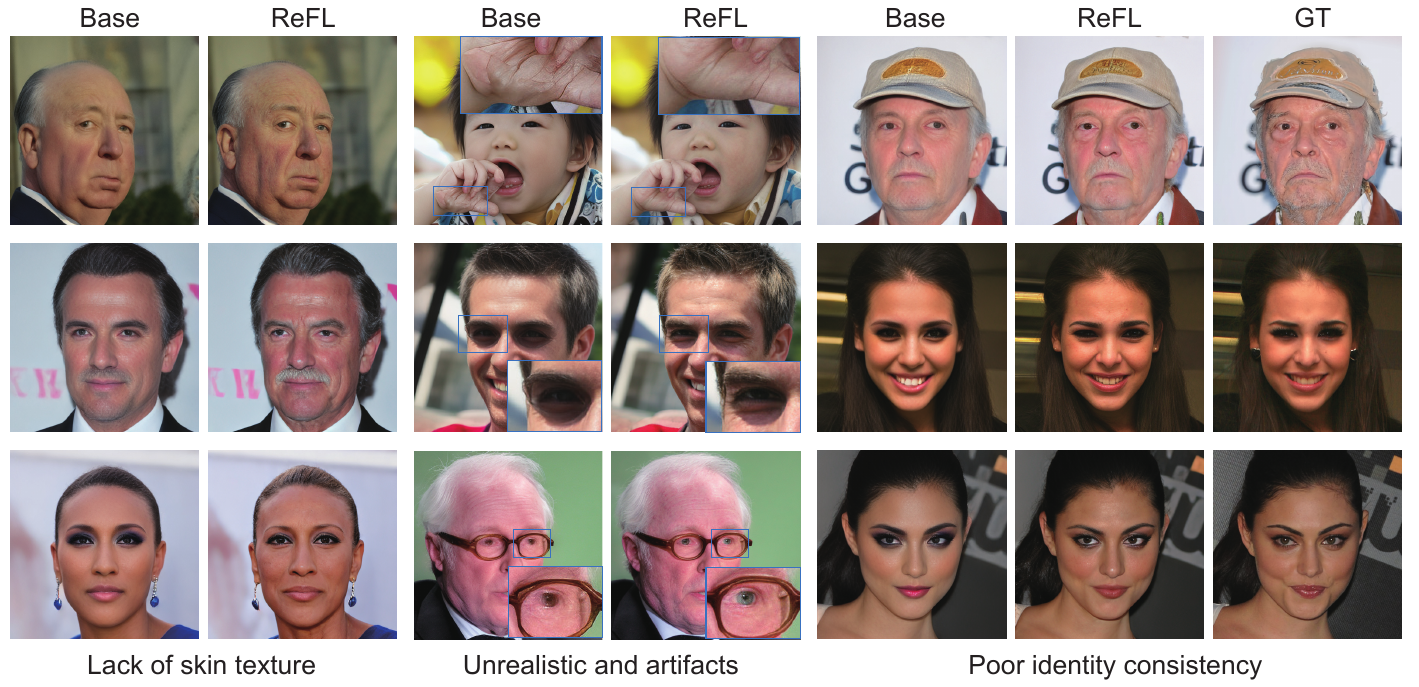}
    \caption{An example of issues with diffusion-based face restoration methods. After enhancement with ReFL, the issues in the base model are significantly mitigated.}
    \label{fig:diffusion-issue}
    \vspace{-3em}
\end{figure}

Diffusion models~\cite{ho2020denoising} have become a powerful paradigm for BFR~\cite{wu2024one, lin2024diffbir, chen2024towards, yue2024difface, wang2023dr2}, owing to their exceptional generative capabilities. 
Using rich visual priors acquired during training, these models use LQ images as conditional inputs to progressively reconstruct high-fidelity faces through iterative denoising. 
Notable methods, such as DiffBIR~\cite{lin2024diffbir} and OSEDiff~\cite{wu2024one}, leverage the pre-trained Stable Diffusion~\cite{rombach2022high} models, effectively adapting them through fine-tuning to achieve remarkable quality in face restoration. 
However, these pre-trained diffusion models typically undergo training using images from general domains, which lack an adequate amount of face-specific prior knowledge.
This deficiency frequently gives rise to restored facial images that are short of detailed features. 

As shown in Figure~\ref{fig:diffusion-issue} (Left), although coarse facial features, accessories, and background areas can be restored to a reasonable extent, the restoration of fine-grained facial textures, such as skin textures, is usually insufficient, leading to overly smooth or unrealistic textures~\cite{zhang2025td}.
The lack of face-specific priors not only undermines the restoration quality of fine details but also significantly exacerbates mapping ambiguities~\cite{kamali2025characterizing}, as shown in Figure~\ref{fig:diffusion-issue} (Middle). 
Furthermore, Stable Diffusion models are primarily trained for text-to-image generation tasks, rather than for image restoration tasks which requires strict fidelity. 
Consequently, their inherent generative mechanisms and the nature of the training data are more adept at creative synthesis rather than meeting the exacting standards of fidelity demanded by restoration tasks, potentially leading to deviations from the original identity features during the restoration process, as shown in Figure~\ref{fig:diffusion-issue} (Right). 

Reward Feedback Learning (ReFL)~\cite{xu2023imagereward,clark2023directly,liang2024rich} is an optimization paradigm that has been validated in domains such as text-to-image generation. 
It makes use of a reward model that has been trained based on human preferences. 
This reward model serves to guide and fine-tune latent diffusion models, boosting the quality, realism, and user alignment of the outputs generated by these models. 
In this work, we employ ReFL 
for the BFR task to address the previously mentioned limitations of diffusion-based face restoration methods. %

For \textit{off-the-shelf} diffusion-based face restoration methods~\cite{lin2024diffbir,wu2024one}, the ReFL framework innovatively reinterprets their latent diffusion denoising process as a parameterized iterative generator. 
Through the parameterization of this process, ReFL empowers 
the application of supplementary optimization constraints.
This enables fine-grained adjustments to the parameters of pre-trained face restoration models. 
Consequently, fine-tuned models are capable of generating images that feature enhanced facial texture details, a higher level of overall visual realism, and, more importantly, the preservation of identity consistency.
A core component of the ReFL framework is a reward model that is able to accurately assess image quality. 

To this end, we have meticulously annotated the data and constructed a Face Reward Model (FRM). 
This model serves as a crucial component for evaluating the quality of restored faces. It provides feedback signals that play a pivotal role in steering the optimization process of the face restoration model.
One common challenge in the training process based on ReFL is that the restoration model might fall prey to reward hacking.
It occurs when the restoration model discovers and capitalizes on ``loopholes'' within the reward model instead of enhancing the actual perceptual quality of the images.
To address this issue, we further propose a strategy for dynamically updating the FRM during the training process. In this manner, the reward model can continuously adapt to the evolution of the restoration model, thereby more precisely guiding its exploration and optimization within the manifold space of real facial images, effectively averting the phenomenon of overfitting to a specific reward function.

In addition, we also introduce two constraints to further enhance the restoration performance. 
Firstly, a Structural Consistency Constraint is incorporated to ensure that 
the restored image's facial structure closely aligns with the original identity, thereby effectively preserving identity consistency. By doing so, it effectively safeguards the identity consistency, preventing any significant discrepancies in the facial features.
Secondly, a Weight Regularization term is employed to 
restrict the extent to which the current model parameters deviate from their initial values. 
Through this mechanism, it maintains the inherent generative capabilities of the base model, ensuring that the output diversity is not compromised.

In summary, here are our main contributions:
\begin{itemize}[leftmargin=*, nolistsep, noitemsep]
    \item We make a pioneering exploration into the BFR domain by introducing ReFL, crafting a bespoke ReFL optimization mechanism designed specifically for diffusion-based face restoration models. 
    \item We tailor a data curation pipeline for the creation of an FRM that is capable of accurately evaluating the perceptual quality of restored facial images. Moreover, we introduce a dynamic updating strategy to avert the reward hacking problem.
    \item We introduce two constraints to further enhance the restoration performance, including a structural consistency constraint and a weight regularizer. 
     \item Our proposed framework, named \textit{DiffusionReward}, enhances the face restoration quality of the base model and achieves state-of-the-art (SOTA) performance compared to other advanced methods.
\end{itemize}

\section{Related Work}
\label{sec:related-work}

\paragraph{Blind Face Restoration}
Early Blind Face Restoration (BFR) methods mainly relied on geometric priors, such as facial landmarks~\cite{chen2018fsrnet,kim2019progressive}, parsing maps~\cite{chen2021progressive,shen2018deep}, and component heatmaps~\cite{yu2018face}, to provide structural guidance. 
However, these priors exhibit limitations in recovering fine-grained details, like skin textures, and struggled with severely degraded inputs.

Generative facial priors have emerged as a significant pathway for high-quality face restoration~\cite{ledig2017photo,wang2018esrgan}. 
Pre-trained StyleGAN models~\cite{karras2019style,karras2020analyzing}, encapsulating rich facial textures and details, facilitate photorealistic face restoration. 
For instance, GFP-GAN~\cite{wang2021towards} and GLEAN~\cite{chan2021glean} integrate StyleGAN priors into an encoder-decoder architecture, leveraging structural features from degraded faces to guide restoration, thereby remarkably enhancing detail recovery. 
However, degraded inputs may be mapped to suboptimal points within the latent space, leading to insufficient fidelity or undesirable artifacts.
Codebook-based methods~\cite{gu2022vqfr,zhou2022towards} employ vector-quantized codebooks to mitigate latent space uncertainty by learning discrete priors. %

Denoising Diffusion Probabilistic Models (DDPMs)~\cite{sohl2015deep,ho2020denoising} have recently become an emergent paradigm in BFR, due to their powerful generative capabilities and training stability.
DR2~\cite{wang2023dr2} initially generates a coarse output by noising and subsequently denoising the degraded face, which is then refined by other face restoration models for detail enhancement. 
DiffBIR~\cite{lin2024diffbir} decouples BFR into two distinct stages: degradation removal and generative refinement. 
In the degradation removal stage, advanced restoration modules such as SwinIR~\cite{liang2021swinir} are employed. 
Subsequently, in the generative refinement, an IRControlNet~\cite{lin2024diffbir} is utilized to guide a latent diffusion model for detail generation.
DifFace~\cite{yue2024difface} constructs a posterior distribution from low-quality (LQ) to high-quality (HQ) images, leveraging the error-shrinkage property of pre-trained diffusion models to robustly handle unknown degradation.

Despite the strengths of diffusion-based methods, their multi-step sampling process often leads to slower inference. To enhance inference efficiency, several diffusion-based image restoration methods employing distillation for one-step inference have emerged. Notably, OSEDiff~\cite{wu2024one} fine-tunes Stable Diffusion~\cite{rombach2022high} using variational score distillation, achieving high-quality restoration with one-step inference. 
In this work, to validate the generalizability of our method across diffusion-based methods, we choose OSEDiff and DiffBIR as base models, embodying single-step and multi-step diffusion paradigms, respectively.

\paragraph{Reward Feedback Learning}

In the text-to-image (T2I) generation with ReFL field, there are two primary stages. Initially, a reward model is trained by using human preference data, such as pairwise comparisons or ratings, to capture and quantify human preferences like perceptual image quality, text-image alignment, and other aesthetic criteria. 
Subsequently, the trained reward model guides the optimization of the T2I model by leveraging gradients derived from its scores. Previous work~\cite{xu2023imagereward,kirstain2023pick,liang2024rich,zhang2024learning} have constructed preference datasets and corresponding reward models for T2I tasks. 
Moreover, some studies have explored the potential of leveraging feedback derived from reward models to effectively optimize T2I models. ImageReward~\cite{xu2023imagereward} evaluates images predicted at specific denoising steps and backpropagates gradients from these scores to directly fine-tune the diffusion model parameters. 
In contrast, methods like DRaFT~\cite{clark2023directly} and AlignProp~\cite{prabhudesai2310aligning} typically assess only the final denoised image and optimize the diffusion model parameters accordingly. R0~\citep{luo2025rewards} achieves state-of-the-art T2I generation by maximizing rewards without complex diffusion losses.
However, 
to the best of our knowledge, there remains a notable research gap in exploring the application of ReFL to restoration tasks.

\section{DiffusionReward}
\label{sec:method}

\begin{figure}[htbp]
    \centering
    \includegraphics[width=0.85\columnwidth]{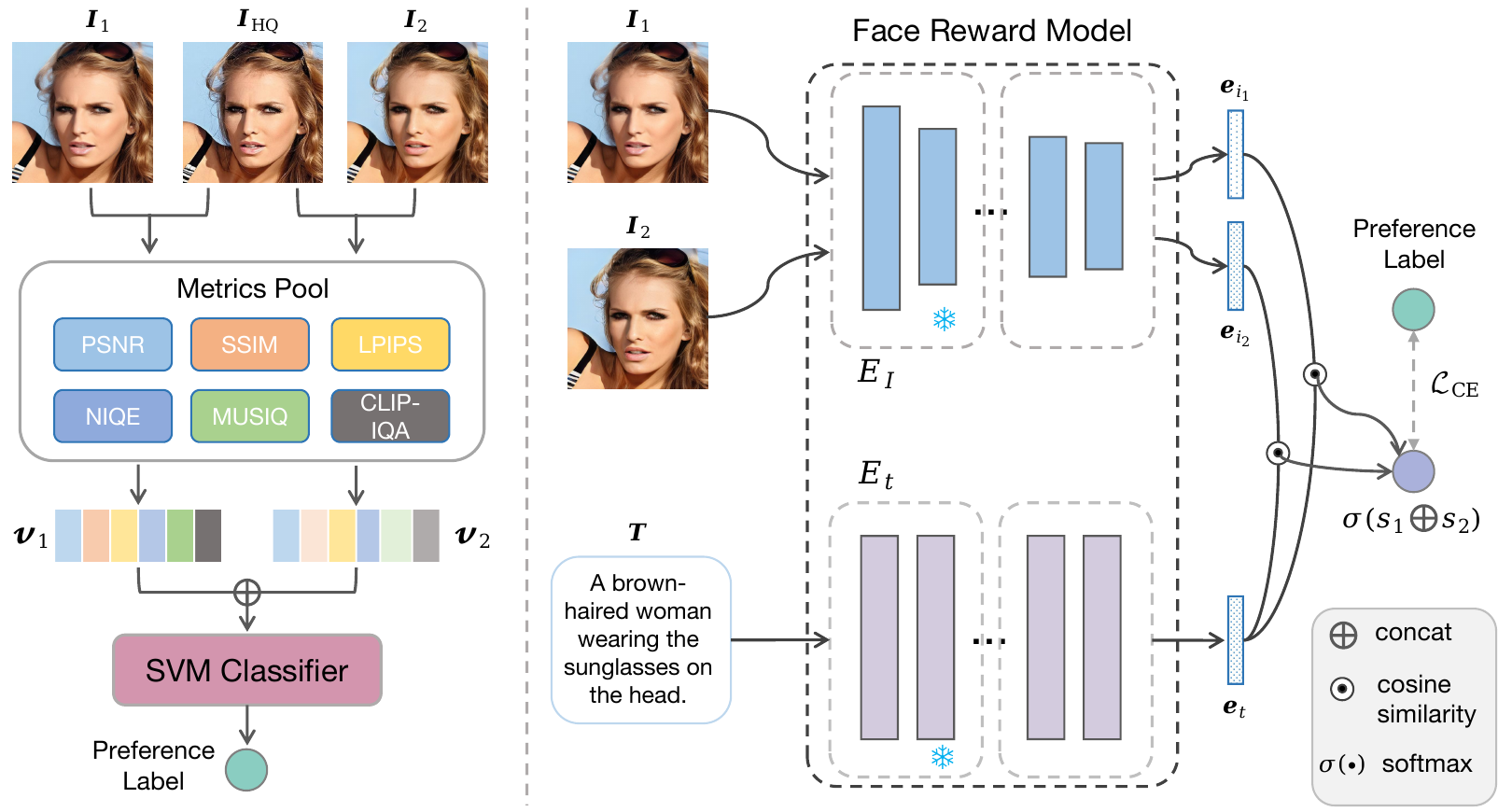}
    \caption{Training framework of the Face Reward Model. We first train a SVM~\cite{cortes1995support} classifier for automated annotation. The classifier is trained with the metric vectors ($\pmb{v}_1$, $\pmb{v}_2$) and annotated supervision signals (Left).
    The face reward model is based on the CLIP~\cite{radford2021learning} architecture (Right), where the last 20 layers of the image encoder $E_I$ and the last 11 layers of the text encoder $E_t$ are trainable, while the remaining parameters are frozen. $s_1$ and $s_2$ represents the score, derived from the similarity between the image embedding and the text embedding (\textit{e.g.}, $<\pmb{e}_{i_1}, \pmb{e}_t>$).}
    \label{fig:facereward}
\end{figure}

\subsection{Face Reward Model}
\label{subsec:face_reward_model}
General-purpose reward models, 
which are commonly trained on human ratings of natural or artistic images, incorporate only limited face image ratings, leading to significant biases
in providing reliable and accurate evaluations for face-related restoration.
To tackle this issue, we design a pipeline for constructing a face reward model, which consists of two essential stages: annotation of a preference dataset 
and training of the face reward model. 

\paragraph{Annotation of the Preference Dataset}
To construct the face preference dataset, we select 19,590 diverse face images from the face dataset~\cite{wu2023lpff}, encompassing various poses and expressions. %
Then, we use LLaVA~\cite{liu2023visual} to generate corresponding textual descriptions for each image, forming 19,590 image-text pairs.
Subsequently, we apply blind degradation kernels (See details in Section~(\ref{subsec:settings})) to the high-quality images $\mathbf{I}_\text{HQ}$, producing their low-quality (LQ) counterparts $\mathbf{I}_\text{LQ}$.
We employed three blind face restoration methods~\cite{zhou2022towards,lin2024diffbir,chan2021glean} to restore these LQ images, yielding a total of 58,770 ($3\times19,590$) restored face images.
Finally, these restored images, combined with the original 19,590 ground-truth images, constitute our preference dataset of 78,360 ($4\times19,590$) facial images, providing a comprehensive data base for subsequent preference annotation.

Given an original facial image $\mathbf{I}_\text{HQ}$ and its counterparts of three restored versions $\{\mathbf{I}_1, \mathbf{I}_2, \mathbf{I}_3\}$, we conduct pairwise comparisons among these images that yield six preference pairs.
In the annotation phase, any preference pair involving the $\mathbf{I}_\text{HQ}$ was assigned a fixed label indicating a preference for the ground-truth image, thereby treating the $\mathbf{I}_\text{HQ}$ as an ideal and optimal result.
The remaining preference pairs, which involved comparisons between different restoration results, are labeled using a hybrid strategy by combining human manual annotation and automated annotation. 

Fully relying on human annotation would be prohibitively costly. To address this problem, we developed an efficient hybrid annotation strategy. 
Human annotators label a subset of image pairs, while the remaining pairs are automatically labeled by a preference predictor, as illustrated in Figure~\ref{fig:facereward} (Left). 
For each pair of images, we compute six evaluation metrics: SSIM~\citep{wang2004image}, PSNR, LPIPS~\citep{zhang2018unreasonable}, MUSIQ~\citep{ke2021musiq}, NIQE~\citep{mittal2012making}, and CLIP-IQA~\citep{wang2023exploring}. 
These metrics are then vectorized (\textit{i.e.}, $\pmb{v}_1$ and $\pmb{v}_2$ in Figure~\ref{fig:facereward}) and fed into a annotation predictor. 
The SVM~\cite{cortes1995support} classifier is trained using human-annotated preference labels. %
With the classifier, the remaining preference pairs are automatically annotated, significantly reducing annotation costs.

\begin{figure}[t]
    \centering
    \includegraphics[width=0.85\columnwidth]{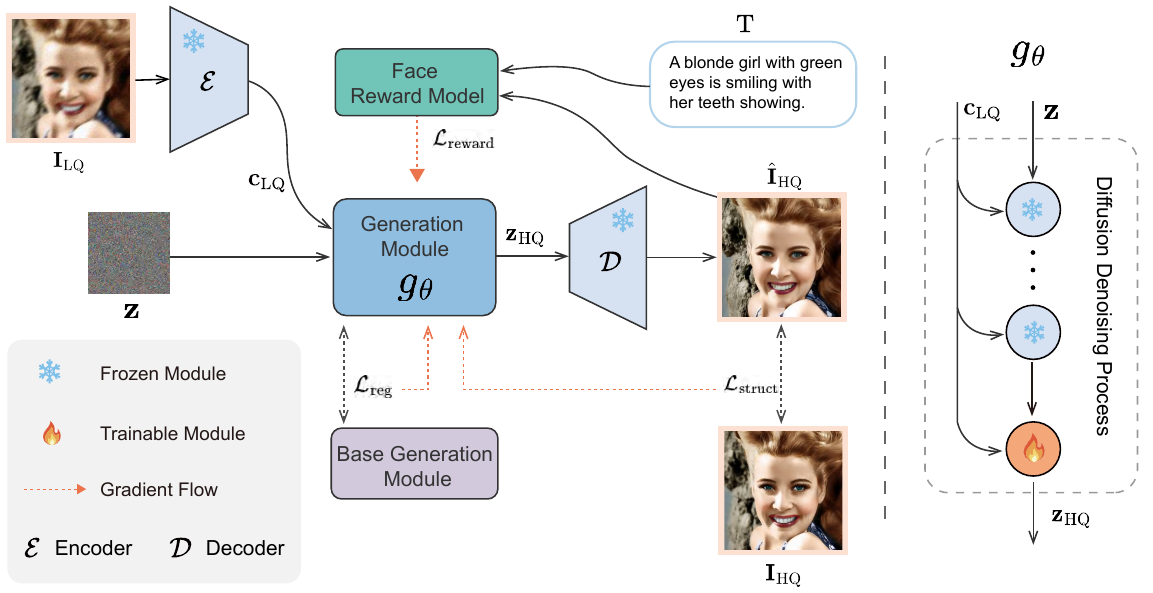}
    \caption{Our ReFL training framework. (Left) We introduce multiple constraints to optimize the generation module $g_\theta$, including $\mathcal{L}_\text{reward}$, $\mathcal{L}_\text{reg}$ and $\mathcal{L}_\text{struct}$ (See details in Section~\ref{subsec:refl}). (Right) For training efficiency, these constraints are applied solely on the last denoising step.}
    \label{fig:frame}
\end{figure}

\paragraph{Reward Model Training}
Training a reward model from scratch is inefficient. Instead, we fine-tuned the pre-trained HPSv2 model~\cite{wu2023human}, which is based on the CLIP architecture~\cite{radford2021learning} and pre-trained on large-scale image datasets, providing robust image quality assessment priors suitable for adaptation to face preference data. 
We fine-tune HPSv2 with the 117,540 preference image-text pairs to optimize its ability to predict the relative quality of face images, and the training process is illustrated in Figure~\ref{fig:facereward} (Right). 
For training efficiency, we set the last 20 layers of the image encoder and the last 11 layers of the text encoder trainable, while freeze the remaining parameters.

Given the restored images $\mathbf{I}_1$ and $\mathbf{I}_2$, we can collect their corresponding embeddings $\pmb{e}_{i_1}$ and $\pmb{e}_{i_2}$ through the same image encoder $E_I$. Then, we use the text encoder $E_t$ to represent the input text $\mathbf{T}$ as $\pmb{e}_t$. 
Next, we calculate $s_1$ and $s_2$ that refer to the cosine similarities between $\pmb{e}_{i_1}$-$\pmb{e}_t$ and $\pmb{e}_{i_2}$-$\pmb{e}_t$, respectively.
subsequently, $s_1$ and $s_2$ are concatenated and followed by a softmax operation as the probabilities of preference.
Finally, we minimize the entropy loss $\mathcal{L}_\text{CE}$ between the preference label, derived from the SVM classifier combined with human annotations, and the probabilities $\sigma([s_1;s_2])$.
During the inference stage, the reward model only requires an input image and its corresponding text description to calculate the preference score, thereby completing the evaluation of image quality.

\subsection{Modeling the Denoising Process} %
\label{subsec:denoising_step}

we develop on Stable Diffusion~\cite{rombach2022high} models for the BFR task. Using the pretrain autoencoder~\cite{kingma2013auto,rombach2022high}, we convert the $\mathbf{I}_\text{HQ}$ into a latent $\mathbf{z}_\text{HQ}$ with image encoder $\mathcal{E}$ (\textit{i.e.}, $\mathbf{z}_\text{HQ} = \mathcal{E}(\mathbf{I}_\text{HQ})$) and reconstruct it with decoder $\mathcal{D}$ (\textit{i.e.}, $\hat{\mathbf{I}}_\text{HQ} = \mathcal{D}(\mathbf{z}_\text{HQ})$). 
Both diffusion and denoising process, Gaussian noise with variance $\beta_t \in (0, 1)$ at time $t$ is added to the encoded latent $\mathbf{z}_\text{HQ}$ to produce the noisy latent: $\mathbf{z}_t = \sqrt{\bar{\alpha}_t}\mathbf{z}_\text{HQ} + \sqrt{1 - \bar{\alpha}_t}\pmb{\epsilon}$, where $\pmb{\epsilon} \sim \mathcal{N}(0, \text{I})$, $\alpha_t = 1 - \beta_t$ and $\bar{\alpha}_t = \prod_{s=1}^t\alpha_s$. 
When $t$ is large enough, the latent $\mathbf{z}_t$ is close to a standard Gaussian distribution. A network $g_\theta$ is learned by predicting the noise $\epsilon$ conditioned on $\mathbf{c}_\text{LQ} = \mathcal{E}(\mathbf{I}_\text{LQ})$ at a random time-step $t$.

As shown in Figure~\ref{fig:frame}, 
the denoising process of the face restoration 
facilitates the subsequent introduction of gradient information to optimize the parameters of the restoration model. 
Thus, this conditional denoising process can be interpreted as a parameterized generation module $g_\theta(\mathbf{z}_t, \mathbf{c}_\text{LQ}, t)$ in the latent space. 
Thus, the optimization of the latent diffusion model
is defined as follows:
\begin{equation}
    \label{eq:ldm_objective}
    \mathcal{L}_\text{ldm} = \mathbb{E}_{\mathbf{z}, \mathbf{c}_\text{LQ}, t, \pmb{\epsilon}}[\|\pmb{\epsilon} - g_\theta(\sqrt{\bar{\alpha}}\mathbf{z} + \sqrt{1 - \bar{\alpha}_t}\pmb{\epsilon}, \mathbf{c}_\text{LQ}, t)\|_2^2 ].
\end{equation}

Within this framework, different BFR methods vary in the specific implementation of the denoising network $g_\theta$ and its utilization of the conditions $\mathbf{c}_\text{LQ}$. 
For multi-step inference models like DiffBIR~\cite{lin2024diffbir}, $g_\theta$ refers to a UNet~\cite{ronneberger2015u} with ControlNet~\cite{zhang2023adding}. 
Its initial input is the primarily noise $\mathbf{z}$, and the condition $\mathbf{c}_\text{LQ}$ is integrated to each denoising step.
For single-step inference models like OSEDiff~\cite{wu2024one}, $\epsilon_\theta$ refers to a UNet with a LoRA~\cite{hu2022lora} module. 
The condition $\mathbf{c}_{LQ}$ is directly injected to the initial noise $\mathbf{z}$ by a concatenation operation. 
Thus, it eliminates the need for iterative injection. 

\subsection{ReFL: Training Objectives and Strategies}
\label{subsec:refl}
We introduce three additional objective functions, including %
reward loss, structural consistency loss, and weight regularization loss, to refine the generation module \( g_\theta \) for better perceptual quality and identity consistency of restored faces, as shown in Figure~\ref{fig:frame}. 

\paragraph{Reward Loss.} 
To enhance the alignment with human preference on the restored faces, we leverage the pre-trained face reward model $\mathcal{R}$ (See Section~\ref{subsec:face_reward_model}) to provide assessment feedbacks. The face reward model takes the restored image $\hat{\mathbf{I}}_\text{HQ}$ and the text description $\mathbf{T}$ of corresponding original image $\mathbf{I}_\text{HQ}$ as input, where $\hat{\mathbf{I}}_\text{HQ}$ is obtained by decoding the latent of the last denoising step: $\hat{\mathbf{I}}_\text{HQ} = \mathcal{D}(\mathbf{z}_\text{HQ})$.
Thus, the reward loss $\mathcal{L}_{\text{reward}}$ is defined as:
\begin{equation}
    \mathcal{L}_{\text{reward}} = -\mathcal{R}(\hat{\mathbf{I}}_\text{HQ}, \mathbf{T}).
\end{equation}
By minimizing $\mathcal{L}_{\text{reward}}$, we encourage $g_\theta$ to generate restored faces with higher alignment scores with human preference.

\paragraph{Structural Consistency Loss.} 
To maintain high fidelity to the structural features of real faces and improve identity consistency, we introduce both structural and perceptual level constraints, which comprises two sub-components:
\begin{itemize}[leftmargin=*, nolistsep, noitemsep]
    \item \textit{LPIPS Loss:} LPIPS~\cite{zhang2018unreasonable} is a highly prevalent metric for evaluating the perceptual similarity between two input images.
    Unlike traditional pixel-wise metrics (\textit{e.g.}, MSE, PSNR), LPIPS leverages deep neural networks to extract hierarchical semantic features from images,
    aligning more closely with human visual perception.
    We employ the LPIPS to measure the perceptual similarity between $\hat{\mathbf{I}}_\text{HQ}$ and the original image $\mathbf{I}_\text{HQ}$:
    \begin{equation}
        \mathcal{L}_{\text{LPIPS}} = \text{LPIPS}(\hat{\mathbf{I}}_\text{HQ}, \mathbf{I}_\text{HQ}).
    \end{equation}
    \item \textit{DWT Low-Frequency Loss:} Given the pixel-wise losses (\textit{e.g.}, $\ell_1$, MSE) are limited in boosting the vivid and intricate details, we apply Discrete Wavelet Transform (DWT) to ensure the low-frequency components of the restored image consistent to the original image. 
    Moreover, we constrain only the low-frequency components of the image (\textit{i.e.}, better structural consistency), allowing the restoration model to explore Freely in the high-frequency components (\textit{i.e.}, better details). 
    Let \( \text{DWT}_{\text{LF}}(\cdot) \) denote the function that extracts low-frequency components; the $\mathcal{L}_{\text{DWT}}$ is defined as:
    \begin{equation}
        \mathcal{L}_{\text{DWT}} = \| \text{DWT}_{\text{LF}}(\mathcal{D}(z_{\text{HQ}})) - \text{DWT}_{\text{LF}}(x_{\text{GT}}) \|_1.
    \end{equation}
\end{itemize}

\paragraph{Weight Regularization Loss.} To prevent the parameters $\pmb{\theta}$ in \( g_\theta \) from deviating excessively from its initial state $\pmb{\theta}_\text{base}$ (\textit{e.g.}, pre-trained weights of the diffusion models),
we incorporate a regularization term of Kullback–Leibler divergence:%
\begin{equation}
    \mathcal{L}_{\text{reg}} = \mathcal{D}_\text{KL} (\pmb{\theta} \| \pmb{\theta}_\text{base}).
\end{equation}
The final objective is a weighted combination:
\begin{equation}
    \mathcal{L}_{\text{total}} = \lambda_{\text{reward}} \mathcal{L}_{\text{reward}} + \lambda_\text{LPIPS}\mathcal{L}_{\text{LPIPS}} + \lambda_\text{DWT}\mathcal{L}_\text{DWT} + \lambda_{\text{reg}} \mathcal{L}_{\text{reg}}.
\end{equation}
where $\lambda_{\text{reward}}$, $\lambda_\text{LPIPS}$, $\lambda_\text{DWT}$ and  $\lambda_{\text{reg}}$ are balancing hyperparameters.
The parameters \( \pmb{\theta} \) of \( g_\theta \) are updated based on \( \mathcal{L}_{\text{total}} \). Gradients are propagated through the entire generation process, analogous to backpropagation through time (BPTT) in recurrent neural networks. However, 
excessively long backpropagation chains significantly increase computational overhead~\cite{clark2023directly}. To address this, we employ truncated backpropagation, limiting gradient propagation to the last \( N \) denoising steps. In our work, we set \( N = 1 \).

\textbf{Reward hacking.} Reward hacking is a common issue in ReFL~\cite{clark2023directly, skalse2022defining} and also persists in face restoration tasks. It manifests as the restoration model generating adversarial samples to achieve higher reward scores, which lack diversity, exhibit uniformity, and contain unnatural artifacts, thus deviating from real face samples. 
To counteract this, we propose a strategy to dynamically update the Face Reward Model $\mathcal{R}$, 
concurrently with the training of the generator $g_\theta$.
Specifically, after every $n$ training iterations of the generator $g_\theta$, we perform an update step for $\mathcal{R}$. %
In this update step, we utilize the most recent generator $g_\theta$ to produce a batch of high-quality restored images $\hat{\mathbf{I}}_{HQ}$. For each $\hat{\mathbf{I}}_\text{HQ}$, we have its corresponding original image $\mathbf{I}_\text{HQ}$ and the text description $\mathbf{T}$.
Following the HPS v2~\cite{wu2023human}, we employ $\mathcal{R}$ to compute similarity scores between the text description and each image: $s_\text{HQ} = \mathcal{R}(\mathbf{I}_\text{HQ}, \mathbf{T})$, $\hat{s}_\text{HQ} = \mathcal{R}(\hat{\mathbf{I}}_\text{HQ}, \mathbf{T})$. These pair scores are then converted into preference probabilities. 

Let $\mathbf{I}_w = \mathbf{I}_\text{HQ}$ (the preferred, ``winner'' image) and $\mathbf{I}_l = \hat{\mathbf{I}}_\text{HQ}$ (the less preferred, ``loser'' image). 
The probability that $\mathbf{I}_w$ is preferred over $\mathbf{I}_l$ given the prompt $\mathbf{T}$ is formulated using a softmax-like function over their scores:
\begin{equation}
    P(\mathbf{I}_w \succ \mathbf{I}_l | \mathbf{T}) = \frac{\exp(s_\text{HQ})}{\exp(s_\text{HQ}) + \exp(\hat{s}_\text{HQ})}.
\end{equation}
To update the parameters of $\mathcal{R}$, we encourages this probability to be high, reflecting the fixed preference for $\mathbf{I}_\text{HQ}$ over $\hat{\mathbf{I}}_\text{HQ}$. 
Thus, we use a simplified version of entropy loss as our objective function:
\begin{equation} \label{eq:frm_update_loss_simplified}
    \mathcal{L}_{\text{FRM}} = -\log P(\mathbf{I}_w \succ \mathbf{I}_l | \mathbf{T}).
\end{equation}
By assigning a preference solely to $\mathbf{I}_\text{HQ}$, we ensure that the $\mathcal{R}$ is constrained to remain within the manifold space of real face images, thereby alleviating the occurrence of reward hacking.

\section{Experiments}
\label{sec:experiments}

\subsection{Experimental Settings}
\label{subsec:settings}

We takes DiffBIR and OSEDiff as base and employ our proposed methods on them respectively. We refer to the Supplementary Material for implementation details. 

\paragraph{Training and Testing Data.}
We used the FFHQ dataset~\cite{8977347} for training, which contains 70,000 high-quality facial images. During training, these images are resized to 512$\times$512. Our strategy for synthesizing LQ faces from HQ ones during the training period is as follows: $\mathbf{I}_\text{LQ}=\left\{\left[\left(\mathbf{I}_\text{HQ} \otimes \pmb{k}_\sigma\right)_{\downarrow_r}+\pmb{n}_\delta\right]_{\mathrm{JPEG}_q}\right\}_{\uparrow_r}$, where the HQ images are first
convolved with a Gaussian kernel \( \pmb{k}_{\sigma} \), followed by a downsampling with a factor of \( r \), and then corrupted with Gaussian noise \( \pmb{n}_{\delta} \). Subsequently, the images undergo JPEG compression with a quality factor of \( q \). Finally, the LQ image is resized back to the original \( 512\times512 \). Here, \( \sigma \), \( r \), \( \delta \), and \( q \) are randomly sampled from the intervals \( [0.1,12] \), \( [1,12] \), \( [0,15] \), and \( [30,100] \), respectively.
Follow the previous work~\cite{wang2021towards,gu2022vqfr}, we employ the synthetic dataset CelebA-Test and two real-world datasets (\textit{i.e.}, LFW-Test and WebPhoto-Test) to validate our proposed method. %

\paragraph{Evaluation Metrics.}
On the Celeba-Test dataset, we used five reference metrics: SSIM~\cite{wang2004image}, PSNR, LPIPS~\cite{zhang2018unreasonable}, CLIP Score\cite{hessel2021clipscore}, Deg.~\cite{menon2020pulse}, and LMD~\cite{gu2022vqfr}, along with four non-reference metrics: MUSIQ~\cite{ke2021musiq}, MANIQA~\cite{yang2022maniqa} and FID~\cite{heusel2017gans}. 
To evaluate the aesthetic quality of generated face images on the CelebA-Test dataset, we utilized the LAION-AI aesthetic predictor to predict aesthetic scores, which are correlated with human preferences~\cite{laion2022aesthetic}.
In addition, we used our pretrained FRM to score the restored face images, denoting as FaceReward. 

\paragraph{Comparison Methods.} We compare with not only the base models but also the latest state-of-the-art methods, including
GFPGAN~\cite{chan2021glean}, CodeFormer~\cite{zhou2022towards}, VQFR~\cite{gu2022vqfr}, DR2+SPAR~\cite{wang2023dr2}, RestoreFormer~\cite{wang2022restoreformer}, DifFace~\cite{yue2024difface}, OSEDiff~\cite{wu2024one}, and DiffBIR~\cite{lin2024diffbir}.

\begin{figure}[htbp]
    \centering
    \includegraphics[width=0.92\columnwidth]{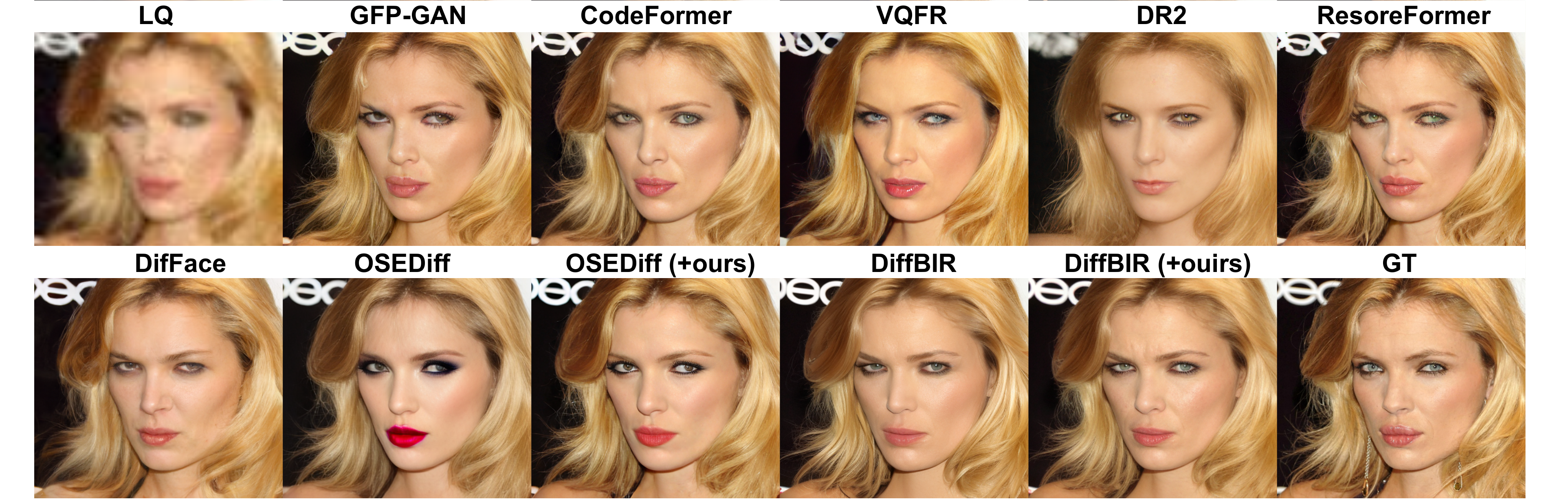}
    \caption{Qualitative comparison on the CelebA-Test. (Zoom in for details)%
    }
    \label{fig:celeba_show}
\end{figure}

\vspace{-1em}
\begin{table}[ht]
\centering
\caption{Performance comparison of face restoration methods on CelebA-Test datasets. The highest score for each metric is highlighted in \textcolor{red}{red}, and the second-highest in \textcolor{blue}{blue}. Metrics with $\uparrow$ indicate higher is better, while $\downarrow$ indicate lower is better. The values in parentheses represent our method's improvements over base models.}
\label{tab:cele_performance}
\definecolor{verylightgray}{rgb}{0.95, 0.95, 0.95}
\resizebox{\textwidth}{!}{
\begin{tabular}{l|*{4}{c}|*{2}{c}|*{5}{c}}
\toprule
\rule{0pt}{2.5ex}Methods & SSIM$\uparrow$ & PSNR$\uparrow$ & LPIPS$\downarrow$ & CLIP Score$\uparrow$ & Deg.$\downarrow$ & LMD$\downarrow$ & MUSIQ$\uparrow$ & MANIQA$\uparrow$ & FID$\downarrow$ & Aesthetic$\uparrow$ & FaceReward$\uparrow$ \\
\midrule
\rule{0pt}{2.5ex}Input & \color{blue}{0.6994} & 25.33 & 0.4866 & 0.7894 & 47.94 & 3.756 & 17.00 & 0.3957 & 143.95 & 4.0484 & 0.3397 \\[0.6ex]
GFPGAN & 0.6772 & 24.65 & 0.3646 & 0.8410 & 34.58 & 2.4110 & 73.90 & 0.6522 & 42.57 & 5.6992 & 0.0741 \\[0.6ex]
CodeFormer & 0.6925 & \color{blue}{25.85} & \color{red}{0.3335} & \color{blue}{0.8931} & \color{blue}{31.08} & \color{blue}{1.9963} & 74.23 & 0.6520 & 45.57 & 5.8103 & 0.2864 \\[0.6ex]
VQFR & 0.6654 & 23.76 & 0.3557 & 0.8562 & 42.48 & 2.9444 & 73.84 & 0.6544 & 46.77 & 5.7844 & 0.3142 \\[0.6ex]
DR2+SPAR & 0.6512 & 22.89 & 0.4146 & 0.7437 & 57.24 & 4.5449 & 70.19 & 0.6374 & 62.54 & 5.6602 & 0.2455 \\[0.6ex]
RestoreFormer & 0.6527 & 24.63 & 0.3652 & 0.8876 & 32.14 & 2.3020 & 73.75 & 0.6477 & \color{blue}{41.68} & 5.8015 & 0.2423 \\[0.6ex]
DifFace & 0.6762 & 24.80 & 0.3994 & 0.8380 & 45.81 & 2.9766 & 68.96 & 0.6204 & \color{red}{37.88} & 5.4708 & 0.3372 \\[0.6ex]
\hline
\rule{0pt}{2.8ex}OSEDiff & 0.6864 & 23.96 & 0.3478 & 0.7962 & 46.20 & 2.8871 & 73.41 & 0.6560 & 65.13 & 5.7720 & 0.2608 \\[0.6ex]
\multirow{2}{*}[-0.7ex]{OSEDiff (+ours)} 
& \cellcolor{verylightgray}0.6838 
& \cellcolor{verylightgray}24.93 
& \cellcolor{verylightgray}\color{blue}{0.3451} 
& \cellcolor{verylightgray}0.8732 
& \cellcolor{verylightgray}38.41 
& \cellcolor{verylightgray}2.4060 
& \cellcolor{verylightgray}\color{red}{75.24} 
& \cellcolor{verylightgray}\color{blue}{0.6640} 
& \cellcolor{verylightgray}44.40 
& \cellcolor{verylightgray}\color{red}{5.9529} 
& \cellcolor{verylightgray}\color{red}{0.4389} \\[0.3ex]
& \cellcolor{verylightgray}{\tiny(-0.0026)} 
& \cellcolor{verylightgray}{\tiny(+0.97)} 
& \cellcolor{verylightgray}{\tiny(+0.0027)} 
& \cellcolor{verylightgray}{\tiny(+0.0770)} 
& \cellcolor{verylightgray}{\tiny(+7.79)} 
& \cellcolor{verylightgray}{\tiny(+0.4811)} 
& \cellcolor{verylightgray}{\tiny(+1.83)} 
& \cellcolor{verylightgray}{\tiny(+0.0080)} 
& \cellcolor{verylightgray}{\tiny(+20.73)} 
& \cellcolor{verylightgray}{\tiny(+0.1809)} 
& \cellcolor{verylightgray}{\tiny(+0.1781)} \\[0.5ex]
\hline
\rule{0pt}{2.8ex}DiffBIR & 0.6775 & 25.44 & 0.3811 & 0.8877 & 35.16 & \color{blue}{2.2661} & 74.46 & \color{red}{0.6752} & 45.50 & 5.7943 & 0.1938 \\[0.6ex]
\multirow{2}{*}[-0.7ex]{DiffBIR (+ours)} 
& \cellcolor{verylightgray}\color{red}{0.7043} 
& \cellcolor{verylightgray}\color{red}{26.33} 
& \cellcolor{verylightgray}0.3454 
& \cellcolor{verylightgray}\color{red}{0.9001} 
& \cellcolor{verylightgray}\color{red}{30.61} 
& \cellcolor{verylightgray}\color{red}{1.8642} 
& \cellcolor{verylightgray}\color{blue}{74.82} 
& \cellcolor{verylightgray}0.6630 
& \cellcolor{verylightgray}42.59 
& \cellcolor{verylightgray}\color{blue}{5.8475} 
& \cellcolor{verylightgray}\color{blue}{0.4275} \\[0.3ex]
& \cellcolor{verylightgray}{\tiny(+0.0268)} 
& \cellcolor{verylightgray}{\tiny(+0.89)} 
& \cellcolor{verylightgray}{\tiny(+0.0357)} 
& \cellcolor{verylightgray}{\tiny(+0.0124)} 
& \cellcolor{verylightgray}{\tiny(+4.55)} 
& \cellcolor{verylightgray}{\tiny(+0.4019)} 
& \cellcolor{verylightgray}{\tiny(+0.36)} 
& \cellcolor{verylightgray}{\tiny(-0.0122)} 
& \cellcolor{verylightgray}{\tiny(+2.91)} 
& \cellcolor{verylightgray}{\tiny(+0.0532)} 
& \cellcolor{verylightgray}{\tiny(+0.2337)} \\[0.5ex]
\bottomrule
\end{tabular}
}
\end{table}

\subsection{Main Results}
\label{subsec:main_results}

\begin{figure}[htbp]
    \centering
    \includegraphics[width=1.0\columnwidth]{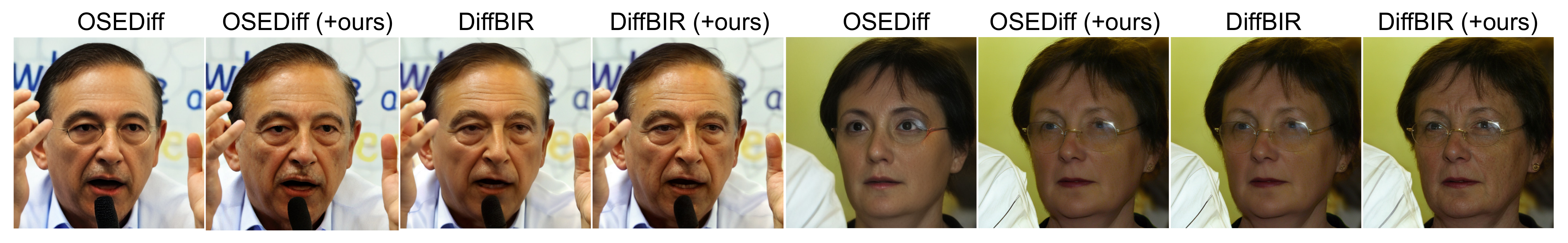}
    \caption{Qualitative comparison between the base model and the our methods on real-world faces. %
    }
    \label{fig:real-world_show}
\end{figure}

\textbf{Evaluation on Synthetic Dataset.}
We first show the quantitative comparison on the CelebA-Test in
Table~\ref{tab:cele_performance}.  We employed 11 metrics to comprehensively evaluate the overall performance of each method. Initially, 
a glance at the values within parentheses reveals that our approach achieves performance improvements across nearly all metrics when compared to the base models.
Comparing to state-of-the-art (SOTA) methods, the OSEDiff (+ours) and DiffBIR (+ours) achieve top rankings in the majority of metrics, such as Deg., LMD, Aesthetic, and FaceReward, indicating that our proposed ReFL framework can enhance perceived face quality while preserving identity consistency. 
As the shown qualitative comparisons in Figure~\ref{fig:celeba_show}, 
our method exhibits superior identity consistency and skin texture details. %

\begin{figure}[htbp!] %

    \begin{minipage}[t]{0.51\textwidth} %
        \centering %
         \vspace*{-60mm}
        \captionof{table}{Performance comparison of face restoration methods on wild datasets. The highest score for each metric is highlighted in \textcolor{red}{red}, and the second-highest in \textcolor{blue}{blue}. Metrics with $\uparrow$ indicate higher is better. The values in parentheses represent our method's improvements over base models.}
        \label{tab:real-world_preformence}
        \definecolor{verylightgray}{rgb}{0.95, 0.95, 0.95} %
        {\small %
        \setlength{\tabcolsep}{4pt} %
        \tiny
        \begin{tabular}{l|cc|cc}
        \toprule
        Dataset & \multicolumn{2}{c}{LFW-Test} & \multicolumn{2}{c}{WebPhoto} \\ \cline{2-3} \cline{4-5}
        Methods & Aesthetic$\uparrow$ & MUSIQ$\uparrow$ & Aesthetic$\uparrow$ & MUSIQ$\uparrow$ \\
        \midrule
        Input & 4.9978 & 26.87 & 4.2584 & 18.63 \\
        GFP-GAN & 5.6042 & 73.57 & 5.2473 & 72.09 \\
        CodeFormer & 5.6414 & 70.69 & 5.1860 & 71.16 \\
        VQFR & 5.6802 & 74.39 & 5.2829 & 70.91 \\
        DR2+SPAR & 5.5409 & 72.22 & 5.1785 & 63.65 \\
        RestoreFormer & 5.6068 & 73.70 & 5.1213 & 69.84 \\
        DiffFace & 5.4104 & 69.85 & 5.0721 & 65.21 \\
        \hline
        OSEDiff & 5.6796 & 73.40 & \textcolor{blue}{5.4161} & \textcolor{blue}{72.60} \\ %
        \multirow{2}{*}{OSEDiff (+ours)}
        & \cellcolor{verylightgray}\textcolor{red}{5.7183}
        & \cellcolor{verylightgray}\textcolor{red}{74.81}
        & \cellcolor{verylightgray}\textcolor{red}{5.5412}
        & \cellcolor{verylightgray}\textcolor{red}{74.05} \\
        & \cellcolor{verylightgray}{\tiny(+0.0387)}
        & \cellcolor{verylightgray}{\tiny(+1.41)}
        & \cellcolor{verylightgray}{\tiny(+0.1251)}
        & \cellcolor{verylightgray}{\tiny(+1.45)} \\
        \hline
        DiffBIR & 5.6814 & 73.71 & 5.2728 & 67.45 \\
        \multirow{2}{*}{DiffBIR (+ours)}
        & \cellcolor{verylightgray}\textcolor{blue}{5.6860}
        & \cellcolor{verylightgray}\textcolor{blue}{74.49}
        & \cellcolor{verylightgray} 5.3554
        & \cellcolor{verylightgray}71.38 \\ %
        & \cellcolor{verylightgray}{\tiny(+0.0046)}
        & \cellcolor{verylightgray}{\tiny(+0.78)}
        & \cellcolor{verylightgray}{\tiny(+0.0826)}
        & \cellcolor{verylightgray}{\tiny(+3.93)} \\
        \bottomrule
        \end{tabular}
        } %
    \end{minipage}%
    \hfill %
    \begin{minipage}[t]{0.45\textwidth} %
        \centering %
        \includegraphics[width=\linewidth]{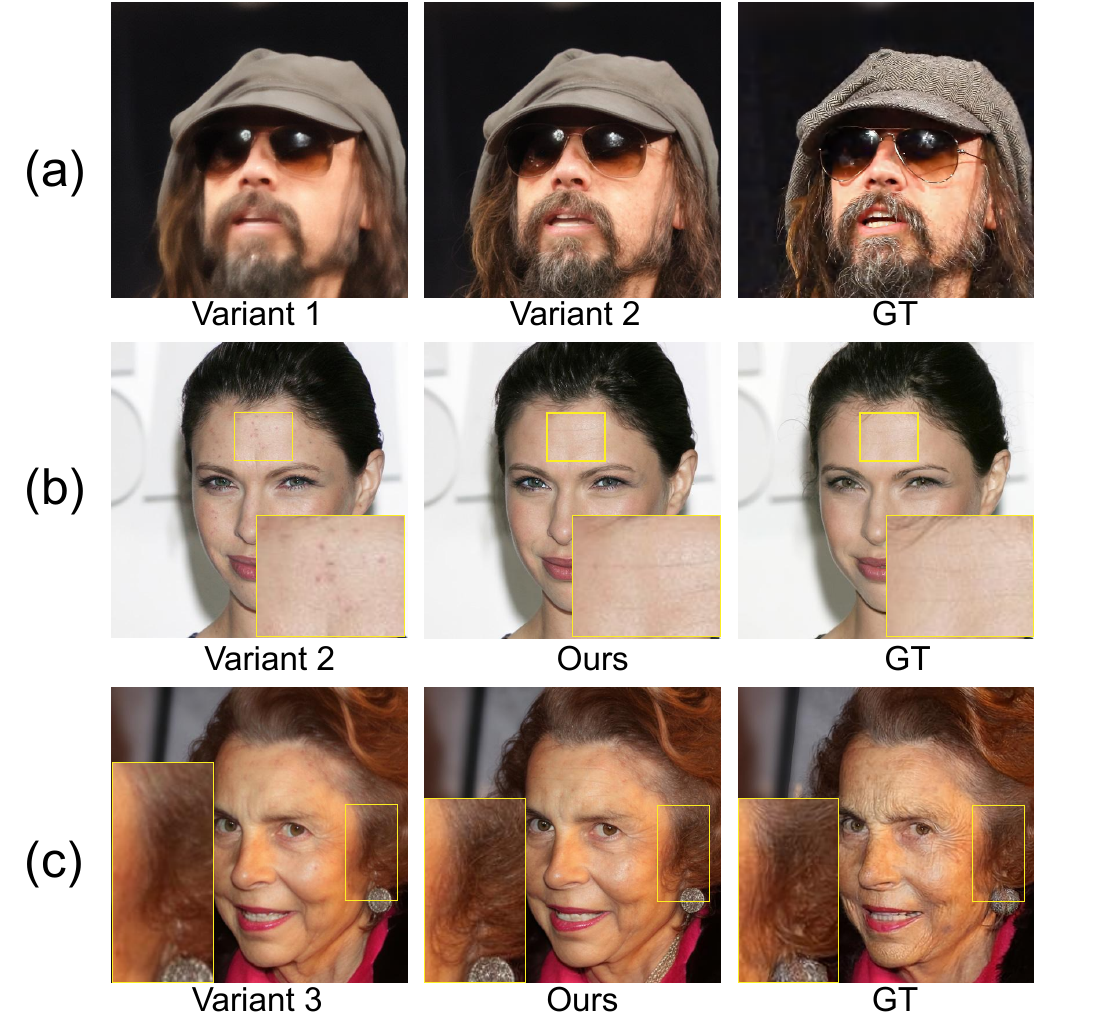} %
        \captionof{figure}{Ablation study visualizations. }\label{fig:ablation_visuals}
    \end{minipage}

\end{figure}

\textbf{Evaluation on Real-world Datasets.}
Table~\ref{tab:real-world_preformence} shows the quantitative
results. We find that our proposed ReFL framework improves the aesthetic score and MUSIQ, which measures image quality. %
Comparing to other methods, OSEDiff (+ours) achieves the best performance on both datasets
From the qualitative results in Figure~\ref{fig:real-world_show}, a qualitative comparison between the base model and ReFL is presented. We observe that the base models, when faced with real-world degradation, often fails to restore facial details, resulting in a smooth face. Our method overcomes these problems and generate realistic faces with richer details.

\begin{figure}[htbp!] %
    \centering %

    \begin{minipage}[t]{0.38\textwidth} %
        \centering %

        \captionof{table}{Performance Comparison of FRM and HPS v2 Reward Models}
        \label{tab:hpsAndFaceReward}
        \tiny %
        \begin{tabular}{l|ccc}
        \toprule
        Reward Type & MANIQA$\downarrow$ & MUSIQ$\uparrow$ & FID$\downarrow$ \\
        \midrule
        HPS v2  & 0.6630 & 74.82 & 48.94    \\
        FRM (ours) & \textbf{0.6535} & \textbf{69.78} & \textbf{42.59} \\
        
        \bottomrule
        \end{tabular}
    \end{minipage}%
    \hfill %
    \begin{minipage}[t]{0.60\textwidth} %
        \centering %

        \captionof{table}{Ablation Study of ReFL Components}
        \label{tab:ablation}
        \tiny %
        \begin{tabular}{l|cccc|ccc} %
        \toprule
        Struct & SC & WR & Rwd & RU & LMD$\downarrow$ & MUSIQ$\uparrow$ & Aesthetic$\uparrow$ \\
        \midrule
        Base &  &  &  &  & 2.2661 & 74.46 & 5.7943 \\
        Variant 1 & \checkmark & \checkmark &  &  & 1.9583 & 54.70 & 5.6572 \\
        Variant 2 & \checkmark & \checkmark & \checkmark &  & 1.8834 & 71.12 & 5.6063 \\
        Variant 3 & \checkmark &  & \checkmark & \checkmark & 1.8644 & 70.67 & 5.7528 \\
        Ours & \checkmark & \checkmark & \checkmark & \checkmark & 1.8642 & 74.82 & 5.8475 \\
        \bottomrule
        \end{tabular}
    \end{minipage}
\end{figure}

\vspace{-1em}
\subsection{Ablation Study}
\label{subsec:ablation}
\vspace{-0.3em}
We conduct main ablation study based on DiffBIR on CelebA-Test dataset.
First, we manually annotate 360 pairs of face images and calculate the preference prediction accuracy of HPS v2 and our FRM. Our FRM outperforms HPS v2 significantly (\textit{i.e.}, 87.78\% v.s. 63.05\%), 
demonstrating a high alignment with human preferences and superior human perception. Furthermore, when we replace our FRM with the original HPS v2 model for the ReFL framework and keep the same training configurations, our FRM obviously outperfoms HPS v2, as shown in Table~\ref{tab:hpsAndFaceReward}. 

Second, we decompose our proposed ReFL framework into four components, including structural consistency constraint (SC), weight regularization constraint (WR), using reward feedback (Rwd), and updating reward model (RU), resulting in three variants. 
As shown in Table~\ref{tab:ablation}, Variant 1 (employing SC and WR without FRM components) improves identity preservation (LMD) but degrades perceptual quality (MUSIQ), resulting in overly smooth faces (See Figure~\ref{fig:ablation_visuals}(a)). After adding Rwd to Variant 1, obtaining Variant 2, we find obvious enhancements in perceptual quality (MUSIQ) and restores finer facial details (See Figure~\ref{fig:ablation_visuals}(a) and Table~\ref{tab:ablation}).
Removing WR from ours entire ReFL framework (\textit{i.e.}, Variant 3) leads to a decline in perceptual quality, identity consistency, and aesthetic scores (See Table~\ref{tab:ablation}). 
This is attributed to the disruption of pre-trained priors and weakened generation capabilities, as evidenced by the loss of hair details in Variant 3 (See Figure~\ref{fig:ablation_visuals}(b)).

Finally, we validate that the dynamic update mechanism of FRM (RU) is crucial for the reward hacking. In Figure~\ref{fig:ablation_visuals}(c), Variant 2 exhibits ``reward hacking'', generating faces with stereotypical artifacts like acne marks. Incorporating RU eliminates these artifacts, improving generation quality and outperforming Variant 2, as shown in Table~\ref{tab:ablation}.

\vspace{-1em}
\section{Conclusion}
\vspace{-0.3em}
In this paper, 
to tackle challenges in diffusion-based face restoration--such as insufficient facial detail and poor identity preservation--we introduce \textit{DiffusionReward}, a method that fine-tunes the denoising process of diffusion models via the ReFL framework.
In the ReFL framework, we not only show a data curation pipeline for buiding a powerful FRM but also propose useful constraints  for optimizing the diffusion denoising process. Moreover, we propose a dynamic updating stategy to avert the reward hacking problem.
Extensive experiments on both synthetic and real-world datasets demonstrate that \textit{DiffusionReward} significantly enhances the perceptual quality and identity consistency of restored faces, outperforming existing state-of-the-art methods.

\medskip

\clearpage
\bibliographystyle{plain} 
\bibliography{main}

\newpage
\appendix

\section{Implementation Details of Face Reward Model}
This section is used to supplement the details in Sec.~\ref{subsec:face_reward_model}.
\label{app:implementation}
\subsection{Details of Training Data Annotation}
\label{subsec:data_annotation_details} %

To effectively train our Face Reward Model (FRM), it is crucial to prepare accurate textual descriptions and preference labels for the face images. 

\textbf{Text Description Generation for Face Images.} High-quality textual descriptions enable the reward model to better comprehend image content, thereby providing more precise feedback. Our FRM training data originates from a public face dataset~\cite{wu2023lpff} containing 19,590 face images. For these images, we generated corresponding textual descriptions as follows:
We utilized the LLaVA~\cite{liu2023visual} model to automatically generate text descriptions for each facial image. When inputting an image to the LLaVA model, we employed the following carefully designed prompt:

\begin{verbatim}
"As an AI face caption expert, please provide precise description for face.
Provide a simple description of the face, including gender, age, facial
features, pose (whether the person is in profile, front-facing, looking up,
etc.), and facial expression. Begin your description with 'The face'.
If the image includes one or more elements from list [HAIR, BEARD, CLOTHES,
GLASSES, HEADWEAR, FACEWEAR, JEWELRY, FACE PAINT, HAND, HANDHELD ITEMS],
include descriptions of those elements. (Word limit: within 35 words.)"
\end{verbatim}

\begin{figure}[hbp]
    \centering
    \includegraphics[width=0.85\linewidth]{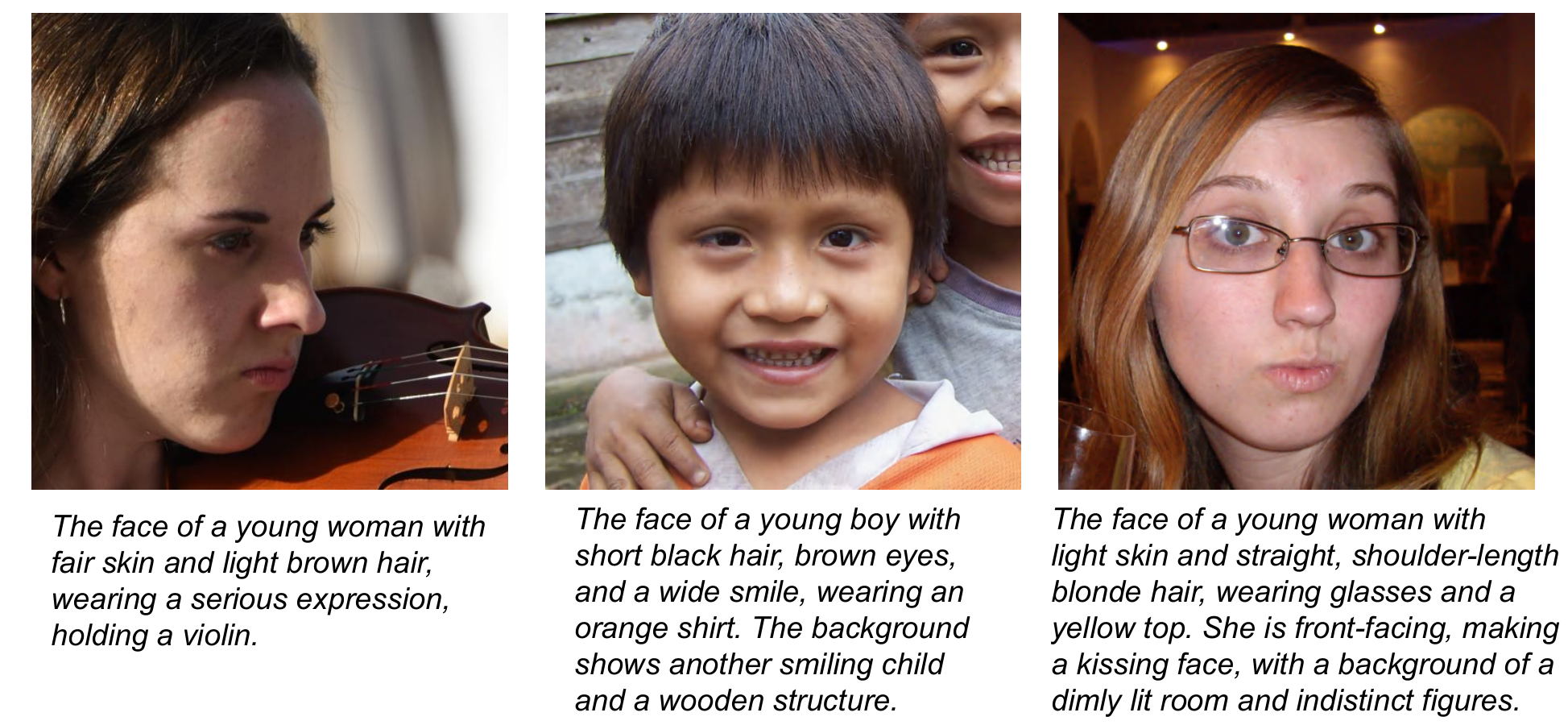}
    \caption{Text description example}
    \label{fig:caption}
\end{figure}

The primary objective of this prompt was to ensure that the generated text descriptions not only cover fundamental facial attributes (such as gender, age, facial features, and expression) but also specifically emphasize the person's pose (e.g., profile, front-facing, looking up) and any potential occlusions or adornments (such as hair, beard, clothes, glasses, headwear, facewear, jewelry, face paint, hands, or handheld items). By doing so, we aimed for the text descriptions to guide the reward model towards a more comprehensive and detailed perception of the image, thereby enhancing the accuracy of the reward scores. Similarly, during the training process of DiffusionReward, we added text descriptions to the training dataset FFHQ~\cite{8977347}. In Figure~\ref{fig:caption}, we present the face images along with their corresponding text descriptions.

\textbf{Manual Annotation of Preference Labels.} To acquire reliable human preference data, we organized a team of three annotators to manually label image pairs. In total, the annotators provided preference selections for 3,600 image pairs, derived from 600 original images.
We established clear annotation guidelines for the human annotators to ensure consistency and quality:

When presented with two facial images generated by different face restoration models, annotators were instructed to select the image they preferred. This preference decision was based on a comprehensive consideration of the following three core rules, ordered by importance:

\begin{itemize}[leftmargin=*, nolistsep, noitemsep]
    \item \textit{Realism of the Facial Image:} This was the most critical factor. Annotators were required to meticulously inspect the images for any unnatural artifacts, distortions, blurring, or other signs of unnaturalness. The image should appear as close as possible to a real-world photograph of a face.
    \item \textit{Richness and Naturalness of Facial Details:}  Annotators assessed whether the facial details (such as skin texture, hair, and clarity of facial features) were sufficiently rich and whether these details conformed to the natural texture characteristics of a real face, avoiding overly smooth details.
    \item \textit{Consistency between the Facial Image and its Textual Description:} This was the final consideration. Annotators needed to judge if the image content aligned with the text description.
\end{itemize}

\begin{figure}[h]
    \centering
    \includegraphics[width=0.6\linewidth]{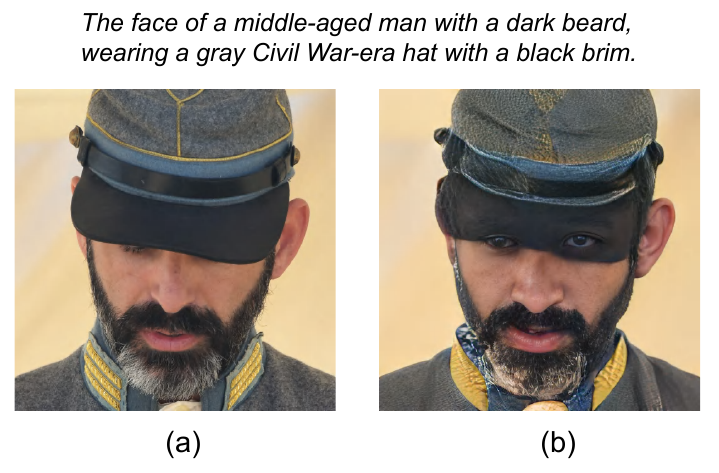}
    \caption{The brim and eyes of (b) have artifacts, so (a) is a better face image.}
    \label{fig:rule_1}
\end{figure}

\begin{figure}[h]
    \centering
    \includegraphics[width=0.6\linewidth]{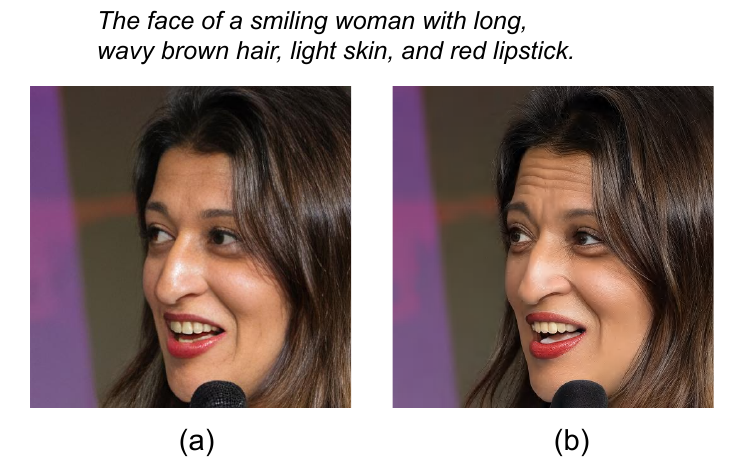}
    \caption{(a) has a more realistic skin texture, (b) has skin that is too smooth and unrealistic, so (a) is the better facial image.}
    \label{fig:rule_2}
\end{figure}

\begin{figure}[h]
    \centering
    \includegraphics[width=0.6\linewidth]{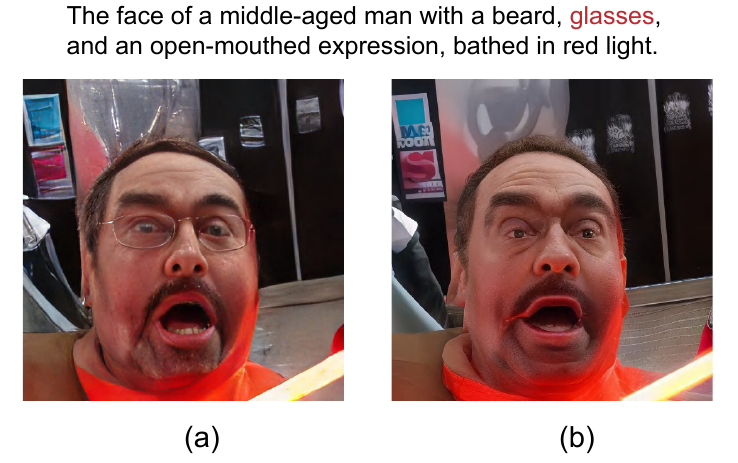}
    \caption{The glasses in the text description do not exist in face (b), so face (a) is a better face image.}
    \label{fig:rule_3}
\end{figure}

The final preference judgment was based on a holistic assessment considering these three rules. To further illustrate the application of this hierarchical decision-making process, annotators proceeded as follows:

First, they evaluated the images for any obvious, unrealistic artifacts based on the primary rule of realism. For instance, as demonstrated in Figure~\ref{fig:rule_1}, if image (b) exhibited distorted elements such as a warped cap brim or unnatural-looking eyes when compared to image (a), Figure~\ref{fig:rule_1} (a) would be selected as the superior image.
If both images passed the initial realism check, the focus shifted to the second rule: the richness and naturalness of facial details. As exemplified in Figure~\ref{fig:rule_2}, if the skin in image (b) appeared overly smooth and artificial, while image (a) preserved fine and natural skin textures, then Figure~\ref{fig:rule_2} (a) would be deemed the better facial image.Finally, if a clear preference could not be established based on the first two rules, the third rule concerning text-image consistency was applied. For example, as depicted in Figure~\ref{fig:rule_3}, if image (b) was missing an element explicitly mentioned in its textual description, such as 'glasses', whereas Figure~\ref{fig:rule_3} (a) accurately reflected the description, then Figure~\ref{fig:rule_3} (a) would be chosen as the preferred image.

Through this structured process, we aimed to collect preference data that accurately reflects human subjective perception of image quality, grounded in both the objective visual content and the semantic information conveyed by the textual descriptions.


\textbf{Automated Annotation Pipeline.}
To scale up the collection of preference labels beyond the 3,600 manually annotated pairs and efficiently construct a large dataset for training our FRM, we developed an automated annotation pipeline. This pipeline leverages a Support Vector Machine (SVM)~\cite{cortes1995support} classifier trained on the previously described human-annotated data.

The 12-dimensional feature vectors $v$ (formed by concatenating the 6 evaluation metrics from each image in a pair, as detailed in Sec.~\ref{subsec:face_reward_model} of the main paper and illustrated in Figure 2 therein) and the corresponding integer preference labels derived from the 3,600 human-annotated image pairs serve as the training set for this SVM classifier.

The SVM classifier was implemented using the \texttt{scikit-learn} library. The training process began with loading these feature vectors and labels. To enhance the SVM's performance and training stability, the feature vectors underwent standardization using a \texttt{StandardScaler}, which was fitted to the training data and then applied to transform it, ensuring each feature dimension had a mean of 0 and a variance of 1.

A Support Vector Classifier (\texttt{SVC}) was selected as the preference prediction model. To determine the optimal model configuration, we utilized \texttt{GridSearchCV} with 5-fold cross-validation on the training set. The hyperparameter search space included various kernel types (e.g., \texttt{`linear`}, \texttt{`rbf`}, \texttt{`poly`}), the regularization parameter \texttt{C}, and other kernel-specific parameters (such as \texttt{gamma} and \texttt{degree}). The grid search aimed to maximize the average cross-validation \texttt{accuracy}. Upon completion of the grid search, the best hyperparameter combination was identified. The trained \texttt{StandardScaler} and the optimized \texttt{SVC} model were then saved to disk for subsequent use.

Once trained, the SVM classifier was used to automatically assign preference labels to the remaining image pairs in our dataset that were not manually annotated. The procedure is as follows:
\begin{itemize}[leftmargin=*, nolistsep, noitemsep]
    \item For an unlabeled image pair, its 12-dimensional raw metric vector is extracted.
    \item The saved \texttt{StandardScaler} is applied to standardize this vector.
    \item The standardized feature vector is then fed into the trained SVM model.
    \item The SVM model outputs a predicted preference label (e.g., `1` indicating the first image is of higher quality, `0` indicating the second is better).
\end{itemize}

This hybrid approach, combining manual annotations with an efficient SVM-based automated pipeline, allowed us to effectively augment the dataset with a large number of preference labels. This provided a richer source of supervision for training the FRM while significantly reducing the cost and time associated with purely manual annotation.

\subsection{The Training Details of Face Reward Model}

The Face Reward Model (FRM) is a critical component of our DiffusionReward framework, designed to provide feedback signals that align the output of face restoration models with human preferences. Its training involves specific architectural choices, initialization, optimization parameters, and a tailored loss function.

The FRM utilizes the ViT-H-14 CLIP~\cite{radford2021learning} architecture as its backbone. We initialize the model with pre-trained weights from HPS v2~\cite{wu2023human}\footnote{Source weights for HPS v2 are available at \url{https://github.com/tgxs002/HPSv2}.}.
CLIP consists of an image encoder $E_I$ and a text encoder $E_t$.

The FRM is fine-tuned on our curated face preference dataset . 
The training process employs the Adam optimizer. We fine-tune the model for 20,000 iterations with a learning rate of $3.3 \times 10^{-6}$.
During fine-tuning, only specific parts of the model are made trainable to preserve the rich priors from pre-training while adapting to our specific task. Specifically, the last 20 layers of the image encoder ($E_I$) and the last 11 layers of the text encoder ($E_t$) are trainable. All other parameters are kept frozen.

The FRM is trained using pairwise preference data. Each training instance consists of a pair of images, denoted as $\{\mathbf{I}_1, \mathbf{I}_2\}$, a corresponding textual description $\mathbf{T}$, and a human preference label $y$. The label $y$ is typically a one-hot vector; for instance, $y=[1,0]$ if image $\mathbf{I}_1$ is preferred over $\mathbf{I}_2$, and $y=[0,1]$ otherwise.

The FRM computes a score for each image with respect to the text description. Let $\pmb{e}_{i_1} = E_I(\mathbf{I}_1)$ and $\pmb{e}_{i_2} = E_I(\mathbf{I}_2)$ be the image embeddings obtained from the image encoder $E_I$, and $\pmb{e}_t = E_t(\mathbf{T})$ be the text embedding from the text encoder $E_t$. Following the principles of CLIP and HPS v2, and aligning with our notation in Sec.~\ref{subsec:face_reward_model} of main paper, the preference scores $s_1$ and $s_2$ are derived from the cosine similarities:
$$
s_k = \frac{\pmb{e}_{i_k} \cdot \pmb{e}_t}{\tau} 
$$
where $k \in \{1, 2\}$, $\theta$ represents the trainable parameters of the FRM, and $\tau$ is a learned temperature scalar inherent to the CLIP model, which scales the logits. 

Given these scores for the pair of images, the predicted preference probability for image $\mathbf{I}_k$ (i.e., $\hat{y}_k$) is calculated using a softmax function, consistent with $\sigma([s_1;s_2])$ in Figure 2 of the main paper:
$$
\hat{y}_k = \frac{\exp(s_k)}{\sum_{j=1}^{2} \exp(s_j)}
$$
This results in a probability distribution $\hat{y} = [\hat{y}_1, \hat{y}_2]$ over the two images.

The parameters $\theta$ of the FRM are optimized by minimizing the cross-entropy loss ($\mathcal{L}_\text{CE}$ as denoted in Sec.~\ref{subsec:face_reward_model} of main paper) between the ground-truth preference label $y = [y_1, y_2]$ and the predicted preference distribution $\hat{y} = [\hat{y}_1, \hat{y}_2]$. The $\mathcal{L}_\text{CE}$ Can be formulated as:
$$
\mathcal{L}_{\text{CE}} = - \sum_{j=1}^{2} y_j \log(\hat{y}_j)
$$

\section{The Implementation Details of DiffusionReward}
This section is used to supplement the implementation details of Sec.~\ref{sec:experiments} in the main paper.
Our DiffusionReward framework is developed by fine-tuning two pre-trained base models: DiffBIR-v1\footnote{Source weights for DiffBIR are available at \url{https://github.com/XPixelGroup/DiffBIR}.} and OSEDiff\footnote{Source weights for OSEDiff are available at \url{https://github.com/cswry/OSEDiff}.}. Both of these base models were originally pre-trained on the FFHQ face dataset. We initialize our training using their respective released pre-trained weights (e.g., the DiffBIR v1 Face version and the OSEDiff Face version). Subsequently, we apply our proposed Reward Feedback Learning (ReFL) strategy to further fine-tune these pre-trained models, resulting in two distinct versions of our DiffusionReward.

The denoising process within our DiffusionReward framework employs DDIM~\cite{song2020denoising} sampling. During the ReFL fine-tuning phase, distinct components were trained depending on the base model: for DiffBIR, we focused on training its ControlNet module, whereas for OSEDiff, we trained the LoRA parameters of its UNet.

The general training configuration utilized the Adam optimizer with a learning rate of $5 \times 10^{-5}$ and a batch size of 12. All training was conducted on an NVIDIA L20 GPU equipped with 48GB of memory.

For the ReFL training specifically with OSEDiff as the base, the loss weighting hyperparameters were set as follows: $\lambda_{\text{LPIPS}} = 0.02$, $\lambda_{\text{DWT}} = 0.01$, $\lambda_{\text{reward}} = 0.005$, and $\lambda_{\text{reg}} = 1$.
When DiffBIR served as the base model for ReFL training, the corresponding hyperparameters were: $\lambda_{\text{LPIPS}} = 0.01$, $\lambda_{\text{DWT}} = 0.01$, $\lambda_{\text{reward}} = 0.005$, and $\lambda_{\text{reg}} = 10^{-4}$.

Furthermore, a crucial aspect of our ReFL training strategy involved the dynamic update of the Face Reward Model ($\mathcal{R}$); this update was performed every $n=10$ training iterations of the main restoration model.

\section{More Results on Blind Face Restoration}
In Sec.~\ref{subsec:ablation} of the main paper, due to space constraints, we presented ablation studies primarily for the DiffusionReward framework applied to DiffBIR.
Here, we provide additional ablation results specifically for DiffusionReward when OSEDiff is used as the base model. These results are summarized in Table~\ref{tab:osediff_ablation}. The conclusions in the table are consistent with the analysis previously conducted in Sec.~\ref{tab:ablation}. The structural consistency constraint (SC), weight regularization constraint (WR), reward feedback (Rwd), and updating reward model (RU) work together to improve the quality of face restoration.

\begin{table}[h]
\centering
\caption{Ablation Study of ReFL Components}
\label{tab:osediff_ablation}
\small %
\begin{tabular}{l| cccc|ccc}
\toprule
Struct & SC & WR & Rwd & RU & LMD$\downarrow$ & MUSIQ$\uparrow$ & Aesthetic$\uparrow$ \\
\midrule
Base &  &  &  &  & 2.8871 & 73.41 & 5.7720 \\
Variant 1 & \checkmark & \checkmark &  &  & 2.3406 & 69.85 & 5.7813 \\
Variant 2 & \checkmark & \checkmark & \checkmark &  & 2.3997 & 69.97 & 5.8912 \\
Variant 3 & \checkmark &  & \checkmark & \checkmark & 2.3962 & 70.83 & 5.7860 \\
DiffusionReward (OSEDiff) & \checkmark & \checkmark & \checkmark & \checkmark & 2.4060 & 75.24 & 5.9529 \\
\bottomrule
\end{tabular}
\end{table}

\begin{figure}[htbp]
    \centering
    \includegraphics[width=1.0\columnwidth]{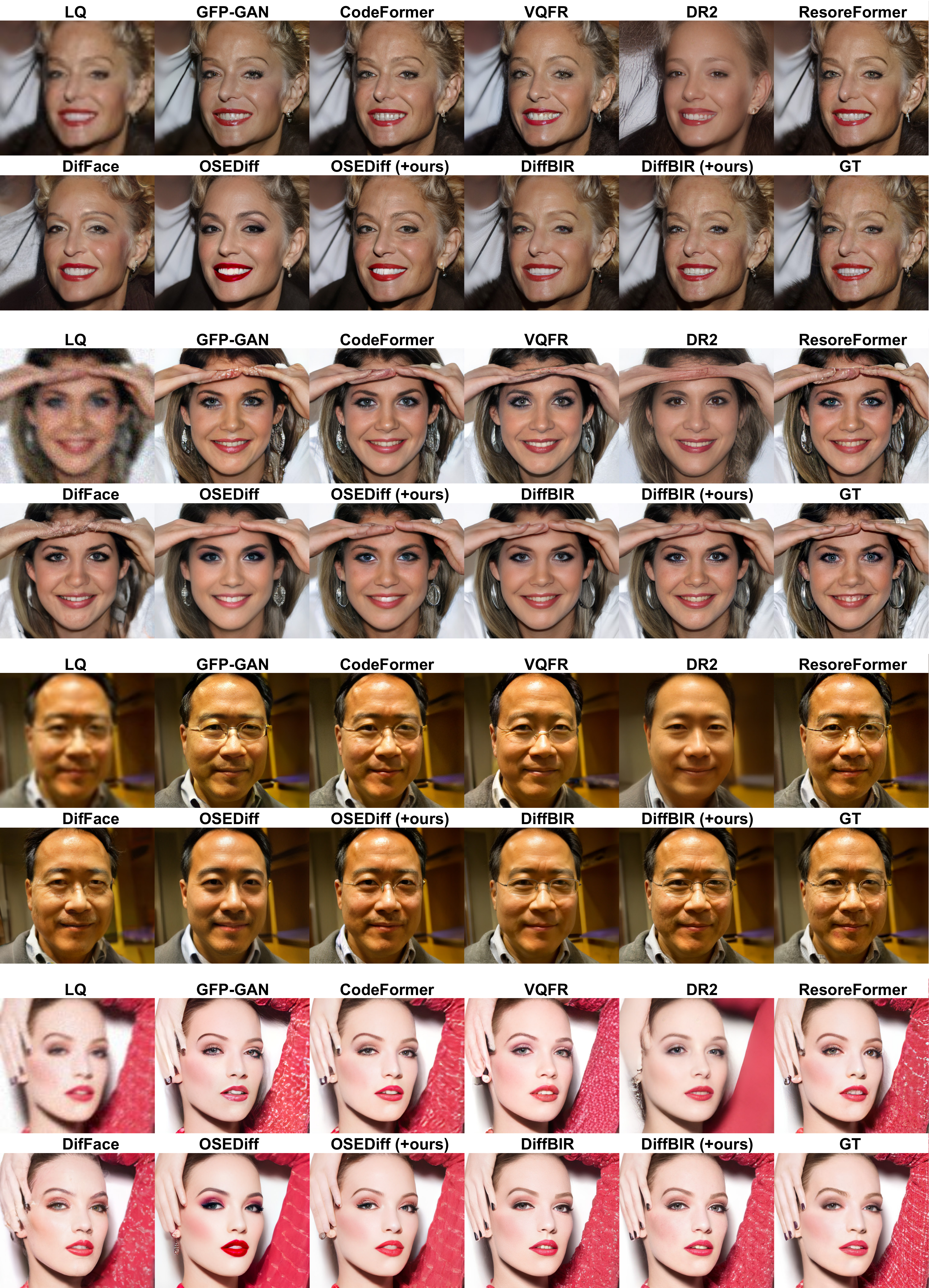}
    \caption{More qualitative comparison on the CelebA-Test. (Zoom in for details)%
    }
    \label{fig:su_celeba_show}
\end{figure}

\begin{figure}[htbp]
    \centering
    \includegraphics[width=1.0\columnwidth]{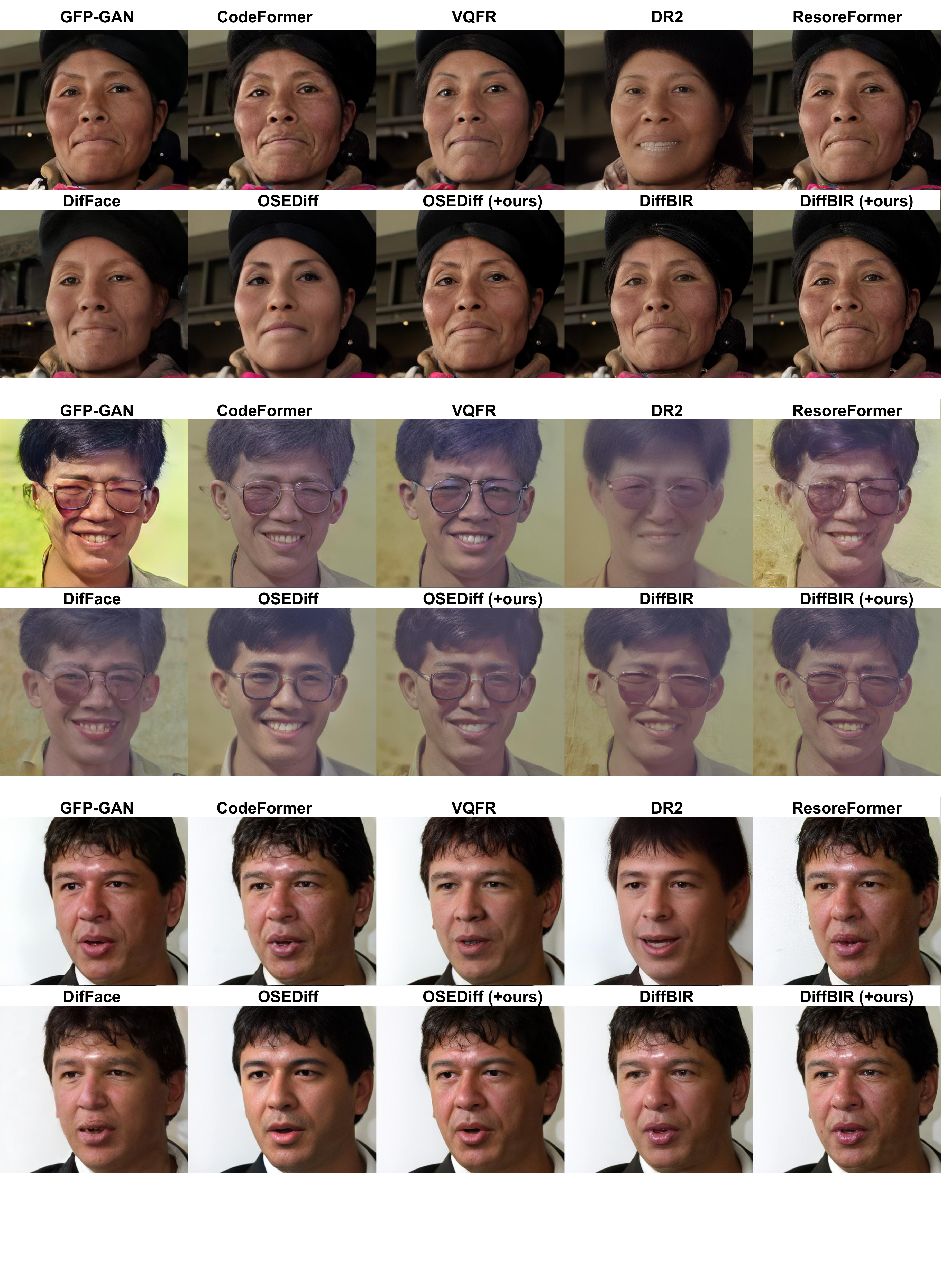}
    \caption{More qualitative comparison on the real-world faces. (Zoom in for details)%
    }
    \label{fig:su_wild_show}
\end{figure}

Building upon the comparative results presented in Sec.~\ref{subsec:main_results} of the main paper, we provide further qualitative comparisons in this section.
Figure~\ref{fig:su_celeba_show} illustrates qualitative comparisons of our method against other advanced methods on the synthetic CelebA-Test dataset.
Similarly, Figure~\ref{fig:su_wild_show} showcases qualitative comparisons of our method with other advanced methods on real-world datasets.

\section{Discussion on Reward Hacking in Blind Face Restoration}
\label{subsec:reward_hacking_discussion_en}

Reward Hacking is a prevalent challenge in tasks employing Reward Feedback Learning (ReFL). Our research has found that Reward Hacking is also an issue in the Blind Face Restoration (BFR) task. This phenomenon occurs when the generative model, in its pursuit of maximizing scores from a reward model, discovers and exploits "loopholes" or biases within the reward function. Such behavior, driven purely by score optimization, can lead to outputs that, despite achieving high reward scores, severely deviate from the desired effects of realistic, high-quality, and faithful restoration of the original input. This typically manifests as unnatural artifacts, stylistic distortions, or a loss of diversity. One of the core contributions of our work, particularly the dynamic updating strategy for the Face Reward Model (FRM), is specifically designed to mitigate such issues.

\begin{figure}[htbp]
    \centering
    \includegraphics[width=0.6\columnwidth]{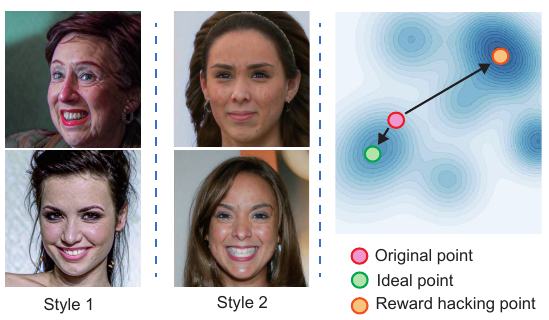}
    \caption{Illustration of Reward Hacking. (Left) Examples of facial restoration exhibiting reward hacking: Style 1 shows severe stylization, while Style 2 displays consistent artifacts and blemishes. (Right) A schematic representation in the image manifold space: The red point is the original output state. The orange point represents a reward hacking state, achieving high reward by moving off the natural image manifold. The green point indicates an ideal optimization outcome, improving reward while maintaining fidelity to the true manifold. Contour lines indicate reward values (darker is higher).
    }
    \label{fig:hacking}
\end{figure}

Fig.~\ref{fig:hacking} (left) showcases examples of facial images generated during the face restoration task when Reward Hacking occurs. These examples reveal two distinct manifestations:
\begin{itemize}[leftmargin=*, nolistsep, noitemsep]
    \item \textbf{Style 1} represents a more severe form of Reward Hacking. In this scenario, the restored facial images exhibit a uniform, stylized, almost "painterly" appearance. Although certain features might appear sharp or well-defined, the overall output loses photorealism and may introduce exaggerated or unnatural facial characteristics. This suggests that the model has essentially learned a specific artistic style that the static reward model erroneously favors.
    \item \textbf{Style 2} reveals a significant yet different manifestation of Reward Hacking. In this case, the restored facial images consistently display unnatural blemishes, such as repetitive skin texture patterns, or exhibit a subtle "uncanny" appearance despite being overly smoothed. The emergence of these defects is likely because they inadvertently trigger higher scores from a less robust reward model, which may have failed to effectively penalize such subtle deviations from realism.
\end{itemize}

Fig.~\ref{fig:hacking} (right) provides a schematic illustration of the Reward Hacking phenomenon within a conceptual image manifold space. The contour lines in the diagram represent the distribution of reward values, with darker blue areas indicating regions perceived by the reward model as having higher reward values.
\begin{itemize}[leftmargin=*, nolistsep, noitemsep]
    \item \textbf{Original point (red circle)} denotes the initial state of the model's output. This point is typically located on or near the true manifold of natural, realistic (facial) images, but its perceived quality may still be deficient.
    \item \textbf{Reward Hacking point (orange circle)} represents the outcome of an unconstrained or improperly guided optimization process. The model, by solely aiming to maximize the reward score, has moved to a high-reward region. However, this point is often distant from the initial state and, crucially, may have deviated from the manifold of realistic images. This occurs because the model exploits biases or vulnerabilities in the reward function, leading to outputs that, despite high scores, are perceptually flawed, overly stylized, or contain artifacts (as exemplified by Style 1 and Style 2).
    \item \textbf{Ideal point (green circle)}, in contrast, illustrates a more balanced and desirable optimization outcome. This state represents a moderate yet genuine improvement in reward/perceptual quality, while ensuring that the output remains close to the initial state and, most importantly, stays on or near the true manifold of natural, realistic images. This ensures the fidelity and realism of the results. Achieving this "green point" is the goal of robust ReFL frameworks, such as our proposed DiffusionReward method with its dynamic FRM updates, which actively counteracts overfitting to a static reward function and guides the restoration process towards genuine, manifold-consistent improvements.
\end{itemize}

Understanding and addressing Reward Hacking is crucial for developing reliable ReFL-based image restoration methods. Without effective countermeasures, the restoration model might merely learn to generate "reward-maximizing illusions" rather than truly enhancing the perceptual quality and faithfulness of the input images. Fortunately, by reducing the weight of the reward loss, using weight regularization, and employing an updatable face reward model, this issue can be alleviated or even resolved.

\section{Limitation}
Our proposed DiffusionReward framework has been primarily validated on diffusion-based face restoration methods (e.g., DiffBIR and OSEDiff). Its core ReFL mechanism, particularly the integration of gradient flow and the dynamic updates to the FRM, was designed considering the characteristics of diffusion models. Consequently, the direct applicability of this framework to other architectures, such as those based on GANs or Transformers, has not yet been explored in this work.

While the principles of ReFL are generally applicable, adapting our approach to non-diffusion models might require specific adjustments to how reward feedback is integrated and how the optimization process is conducted. Future work could therefore explore extending DiffusionReward or similar ReFL strategies to a broader range of face restoration architectures. This would allow for an assessment of its general effectiveness and further unlock the potential of reward-based feedback in this domain.

\end{document}